**ORIGINAL ARTICLE**

# Damage detection using in-domain and cross-domain transfer learning

Zaharah A. Bukhsh[1] · Nils Jansen[2] · Aaqib Saeed[1]

**Abstract**
We investigate the capabilities of transfer learning in the area of structural health monitoring. In particular, we are interested in damage detection for concrete structures. Typical image datasets for such problems are relatively small, calling for the transfer of learned representation from a related large-scale dataset. Past efforts of damage detection using images have mainly considered cross-domain transfer learning approaches using pre-trained IMAGENET models that are subsequently fine-tuned for the target task. However, there are rising concerns about the generalizability of IMAGENET representations for specific target domains, such as for visual inspection and medical imaging. We, therefore, evaluate a combination of in-domain and cross-domain transfer learning strategies for damage detection in bridges. We perform comprehensive comparisons to study the impact of cross-domain and in-domain transfer, with various initialization strategies, using six publicly available visual inspection datasets. The pre-trained models are also evaluated for their ability to cope with the extremely low-data regime. We show that the combination of cross-domain and in-domain transfer persistently shows superior performance specially with tiny datasets. Likewise, we also provide visual explanations of predictive models to enable algorithmic transparency and provide insights to experts about the intrinsic decision logic of typically black-box deep models.

**Keywords** Damage detection · Transfer learning · Pre-trained models · In-domain learning · Cross-domain learning · Visual inspection

## 1 Introduction

Civil structures such as bridges are reaching their end of service life due to aging, increased usage, and adverse climate impact [14]. Asset owners have to employ human experts to conduct periodic visual inspections to ensure structural safety and usability. In particular, so-called condition scorecards are used to rate the condition of bridges, whereas damage details are captured in images for further analysis [25]. Given that countries have thousands of bridges, for instance, 60,000 bridges across the USA and 39,000 bridges in Germany, such expert-driven visual inspections are very costly and labor-intensive.

Toward a more automated process, unmanned aerial vehicles (UAVs) equipped with cameras, thermal infrared sensors, GPS, as well as light detection and ranging (LiDAR) sensors have proved useful to characterize cracks, spalls, concrete degradation, and corrosion [13, 36, 44]. Although UAVs are cost-effective and safe for harder-to-reach areas, they capture both damaged and structurally sound parts of bridges. As a consequence, an enormous amount of high-dimensional sensory data is created [9]. Manual damage identification from such an enormous data source demands tremendous efforts and is prone to discrepancies due to human errors, fatigue, and poor judgments of bridge inspectors [2, 46]. For image data, the involved subjectivity in a visual inspection process results in inaccurate outcomes and poses serious concerns for public safety [40], as shown by the collapsing incidents like the Malahide viaduct [50] or the I-35 Minneapolis bridge [63].

✉ Zaharah A. Bukhsh
z.bukhsh@tue.nl

Nils Jansen
n.jansen@science.ru.nl

Aaqib Saeed
a.saeed@tue.nl

[1] Eindhoven University of Technology, Eindhoven, The Netherlands

[2] Radboud University, Nijmegen, The Netherlands

The large unlabeled datasets from visual inspections call for state-of-the-art machine learning methods. In particular, we focus on damage detection of concrete surfaces. The field of computer vision has witnessed unprecedented developments since the advent of deep learning and the availability of large-scale annotated datasets, such as IMAGENET. Deep neural networks, specifically convolutional neural networks (CNNs), have shown strong performance in tasks such as object detection [20, 21, 53], image segmentation [29], image synthesis [47], and reconstruction [22], with promising applications in multiple domains.

While the use of deep learning techniques for structural health monitoring of civil structures is gaining momentum [18, 70],
to the best of our knowledge only six bridge inspection datasets [26, 28, 37, 11, 42, 67] are publicly available that can be utilized to develop automated damage detection models. These datasets are relatively small scale because of the tedious process of image acquisition and data labeling by experts.

Due to the limitation of small-scale datasets, recent studies for damage detection of bridges, tunnels, and roads have adopted transfer learning as the de facto standard for crack detection [7, 31, 52], pothole identification [39, 68], and related defect classification tasks [67]. These studies exclusively rely on cross-domain transfer learning, where IMAGENET is used as the source dataset (or upstream task), which is then fine-tuned for specific downstream tasks. However, there is skepticism about employing transfer learning for disparate target domains [19, 23, 27, 32]. For instance, [24] showed that reusing of IMAGENET features offers little benefit to performance when evaluated with medical imaging. Similar to medical images, bridge inspection datasets are fundamentally different from the IMAGENET dataset in terms of the number of classes, the quality of images, the size of the areas of interest, and the task specification.

The problem of choosing the relevant source dataset to learn general-purpose representations for the visual inspection target dataset is overlooked in the literature. Recent studies on damage detection adopt IMAGENET dataset as a default choice of source dataset, which falls under the realm of cross-domain transfer learning. This work seeks to eliminate this literature gap and performs a fine-grained investigation to study the potential advantages and downsides of *cross-domain transfer learning for structural damage detection*. We also learn and transfer *in-domain representation* and *its combination with cross-domain transfer strategies* for automated damage detection tasks. Specifically, in-domain transfer learning refers to a strategy in which the source and target dataset belong to a similar domain. Furthermore, we evaluate the transferability of learned representations under a low-data regime.

Besides transfer learning strategies, we also compared three state-of-the-art pre-trained models (InceptionV3, VGG16, and ResNet50) to small vanilla CNN models for the six publicly available bridge inspection datasets. The models are compared and evaluated using different initialization techniques, namely random, in-domain, and cross-domain initialization.

This work offers following contributions and findings:

- We conduct rigorous evaluations of transfer strategies with different initialization settings and network architectures for damage detection using the six publicly available visual inspection datasets. We also establish benchmarks for publicly available damage detection datasets.
- We develop multiple CNN models using cross-domain and in-domain transfer learning strategies for damage detection (classification) tasks.
- We find that the cross-domain transfer of pre-trained models proves helpful only when they are fine-tuned on the target task.
- We show that the smaller and simpler CNN models provide comparable performance to standard architectures when used as a fixed-feature extractor.
- We note that in-domain representations provide performance improvements compared to the random initialization.
- We show that the in-domain representations together with IMAGENET show performance gains compared to an independent transfer setting, particularly when the target dataset has fewer training samples.
- We provide visual explanations of multilayered deep neural networks to reveal their learning mechanisms. Such an interpretability analysis enables algorithmic transparency, validates the robustness of trained models, and gains practitioners' trust in typically black-box models.

The paper is structured as follows: Section 2 first introduces standard IMAGENET models, and then concerns related work about damage detection using deep transfer learning and feature transfer using pre-trained models. Section 3 introduces the publicly available bridge inspection datasets. Section 4 explains the methodology of our approach, along with the experimental setup. Section 5 presents the results of the experiments and provides comparisons to the baselines. An interpretability analysis of the developed models is given in Section 6. Finally, the conclusions of this study are presented in Section 7.

## 2 Background and related work

This section first provides a brief introduction to pre-trained models and their architectures for transfer learning. Next, we introduce related studies about automatic damage detection using deep (transfer) learning. Then, we highlight studies about possible limitations of generic feature transfer using pre-trained models. Additionally, for completeness, we highlight related topics from deep learning domains that are often associated with transfer learning.

### 2.1 Pre-trained models and standard neural architectures

Deep convolutional neural networks (CNNs) are a special type of multilayered neural network that have contributed to spectacular advancements in the computer vision field. Even though CNNs have been used for object detection tasks for several years [35], their widespread recognition and adoption are relatively recent, when the AlexNet model achieved state-of-the-art performance in the IMAGENET classification challenge [33]. Owing to the AlexNet success, the practice of training a CNN on IMAGENET (i.e., pre-training) and then adapting (e.g., fine-tuning) it to a target task has become a norm for several computer vision problems. Since pre-training on a large dataset is computationally expensive and time-consuming, several pre-trained models have been made publicly available by academics and industry.

In this study, we evaluate three state-of-the-art deep CNN models, namely VGG16 [55], Inception-v3 [58], and ResNet50 [23], which are lately being employed for damage detection tasks in structural health monitoring. An in-depth explanation of CNNs is beyond the scope of this study. The interested reader may refer to [65]. The section provides an architectural overview of the selected pre-trained models.

VGG16 invented by Visual Geometry Group of Oxford University [55] won the first place for IMAGENET object localization and the second place in the image classification challenge in 2014. Figure 1 provides a schematic representation of the VGG16 network. VGG16 improves over the AlexNet model by introducing a uniform structure of five convolution blocks having a fixed filter size of 3x3 with the stride of one. With its 13 convolutional layers and three fully connected layers, VGG16 demonstrated that the deeper models are desirable for improved classification accuracy. However, the VGG's simple architecture comes at a high computational cost, with 138 million parameters (3x more than AlexNet) and prolonged training time.

Inception-v3 improves the GoogLeNet model that won the 2014's IMAGENET classification challenge. In addition to classification accuracy, Inception-v3 focuses on reducing the computational expense, improving the training efficiency, and optimizing the network for easier adaption to different use cases [59]. It is one of the first non-sequential models that emphasize a wider and deeper network by employing an *inception module*, as illustrated in Fig. 2. Instead of applying a single convolution with a fixed-size kernel, multiple convolutions with varying kernels (e.g., factorized 7x7) are applied to an input layer, which are then concatenated for another inception module. Additionally, to avoid the gradient vanishing problem, two auxiliary classifiers with batch normalization are included. Inception-v3 is 48 layers deep with 11 concatenated inception modules.

ResNet50 was the winner of famous 2015's IMAGENET classification challenge. Residual network with several variants (such as ResNet10 and ResNet52) mainly dealt with the vanishing gradient problem. The residual network introduced an intuitive idea of skip connections with identity function mappings, as shown in Fig. 3. Specifically, the residual block learns the residual mappings, which captures the difference between actual input and approximate output rather than fitting the desired underlying mappings [24]. By approximating the identity function, residual learning does not introduce additional parameters and keeps the computational complexity similar to deep vanilla networks. ResNet50 consists of four stages, each containing two to six residual blocks with a kernel size of 1x1 and 3x3. In progressing from one stage to another, the channel width is doubled while the input size is reduced to half.

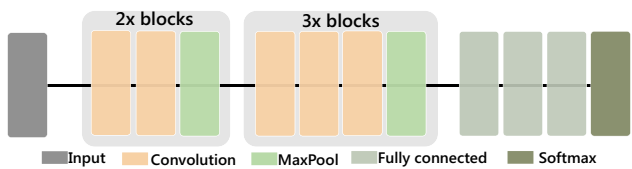

**Fig. 1** Schematic representation of VGG16

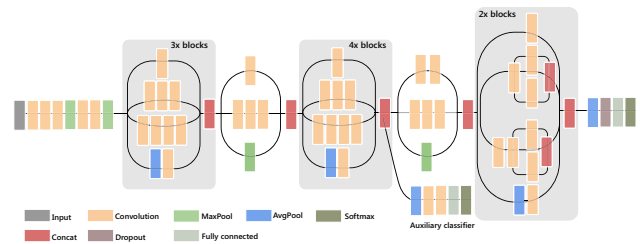

**Fig. 2** Schematic representation of Inception-v3

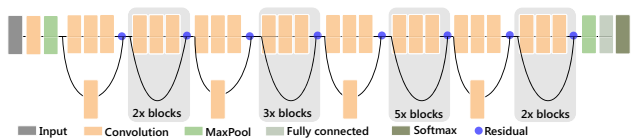

**Fig. 3** Schematic representation of ResNet50

Various terminologies are interchangeably used to refer to the pre-trained models. The CNN architectures are referred to as *pre-trained models* or IMAGENET *models*, when pre-trained using IMAGENET weights. These networks without learned representations are referred as *standard architectures* due to their wide usage in multiple application domains.

## 2.2 Damage detection using deep (transfer) learning

Visual inspection datasets are small scale and are expensive to curate. Therefore, several studies adopt standard architectures—such as VGG16, Inception-v3, LeNet, YOLO, or ResNet50 often pre-trained with IMAGENET representations—for the detection of cracks [31, 56, 61, 75], potholes [39], spalls [68], and multiple other damages including corrosion, seapage, and exposed bars [17, 28, 72, 78]. In Table 1, we provide a non-exhaustive list of recent literature studies that have used transfer learning for concrete damage detection tasks. It is notable that except for one, all studies have utilized IMAGENET as a source dataset. Additionally, pre-trained CNN models are also being utilized for vibration-based damage localization [1, 4, 69], condition assessment [30], and fault diagnosis [66].

Specific (visual) detection tasks in the damage recognition setting are strongly influenced by various operating conditions, such as surface reflectance, roughness, concrete materials, coatings, and weather phenomena for different components of a bridge [42]. Given the unique characteristics of visual inspection dataset, the problem of choosing an appropriate source dataset for damage detection remained an open question. This study seeks to address this by conducting a thorough investigation to evaluate the transferability of cross-domain representations (i.e., from IMAGENET) for damage detection tasks.

**Table 1** List of recent studies that utilize transfer learning for concrete damage detection

| Proposed method | Year | Task | Pre-trained model | Source dataset |
| --- | --- | --- | --- | --- |
| CovNet Model [56] | 2017 | Concrete surface cracks | VGG16 | IMAGENET |
| Custom CNN [3] | 2017 | Corrosion detection | VGG16 | IMAGENET |
| CNN & edge detection [10] | 2018 | Crack detection in concrete | AlexNet | IMAGENET |
| Custom CNN [11] | 2018 | Crack detection in concrete | AlexNet | IMAGENET |
| Custom CNN [39] | 2018 | Road damage detection | MobiNet | IMAGENET |
| TERNAUSNET [5] | 2019 | Crack segmentation | VGG16 | IMAGENET |
| Custom CNN [17] | 2019 | Structural damage detection | Inception-v3 | IMAGENET |
| Custom CNN [78] | 2019 | Bridge defects detection | Inception-v3 | IMAGENET |
| Multi-classifier [28] | 2019 | Concrete bridge defects | Inception-v3 | IMAGENET |
| Custom CNN [12] | 2019 | Crack detection in concrete | VGG16 | IMAGENET |
| Semantic seg. network [75] | 2019 | Crack detection in concrete | VGG16 | IMAGENET |
| Custom CNN [64] | 2020 | Concrete surface roughness | ResNet | IMAGENET |
| Extension of YOLOv3 [73] | 2020 | Bridge damage detection | YOLOv3 | COCO |
| Improved EfficientNetB0 [57] | 2020 | Crack detection in concrete | MobileNetV2 | IMAGENET |
| Modifed YOLO [61] | 2021 | Crack detection in concrete | YOLO | IMAGENET |
| Shallow CNN [31] | 2021 | Surface crack detection | LeNet-5 | IMAGENET |
| Custom CNN [52] | 2021 | Road surface damage detection | ResNet34 | IMAGENET |
| CNN for pixel level [7] | 2021 | Crack classification & seg. | Multiple models | IMAGENET |

## 2.3 Feature transfer using pre-trained models

Transfer learning via IMAGENET representations is a widely adopted standard for learning from small-scale datasets across several domains. The availability of a multitude of pre-trained models such as MoBiNet, YOLO, ResNet, or GoogLeNet has further encouraged the application of transfer learning beyond the standard datasets consisting of day-to-day objects. The learned representations are typically utilized either via a fixed-feature extraction or fine-tuning methods depending on the target tasks.

Due to the widespread adoption of transfer learning, several fine-grained studies examine the transferability of features concerning the number of layers, the order of fine-tuning [38, 60, 71, 74], the generalizability of the architecture or learned weights [32, 48], and the characteristics of data used for pre-training [27]. Besides the popularity of transfer learning and its perceived usefulness, it has been argued that feature transfer does not necessarily deliver performance improvements compared to learning from scratch [23, 49]. Moreover, learned representations from pre-trained models may be less generic than often suggested [32], and the choice of data for pre-training is not as critical for the transfer performance [27].

## 2.4 Related topic

In the context of in-domain and cross-domain strategies, domain adaption is often misinterpreted as a general transfer learning method. In fact, domain adaption is a sub-category of transfer learning in which both target and source datasets are required during training, and the target dataset is weakly labeled or unlabeled [45]. However, in a standard transfer learning setting, the target dataset is supervised (labeled), and only pre-trained models can be used for learning without explicit access to the source dataset. Further details about transfer learning and proposed strategies are provided in the following sections.

## 3 Datasets

Due to the ubiquitous nature of images, several traditional application domains such as retail, automotive, or agriculture have benefited from massive labeled datasets to develop deep neural networks to solve various tasks. These natural datasets are relatively easy to label and do not require specific domain expertise. On the other hand, bridge inspection datasets are very scarce due to the vital domain knowledge required to identify and label damages. Moreover, the differences in human expertise for annotation make these datasets susceptible to noisy labels. To the best of our knowledge, only six visual inspection datasets are publicly available. Datasets having less than 100 images are too small to train and evaluate CNNs and are therefore not considered here. Table 2 provides a brief overview of the datasets. A few typical example images are shown in Fig. 4.

Table 2 Overview of (bridge) visual inspection datasets

| Dataset | Instances | Classes | Problem |
| --- | --- | --- | --- |
| CDS [26] | 1,027 | 2 | Binary |
| SDNETv1 [11] | 13,620 | 2 | Binary |
| BCD [67] | 5,390 | 2 | Binary |
| ICCD [37] | 60,010 | 2 | Binary |
| MCDS [28] | 2,411 | 10 | Multi-label |
| CODEBRIM [42] | 8,304 | 6 | Multi-label |

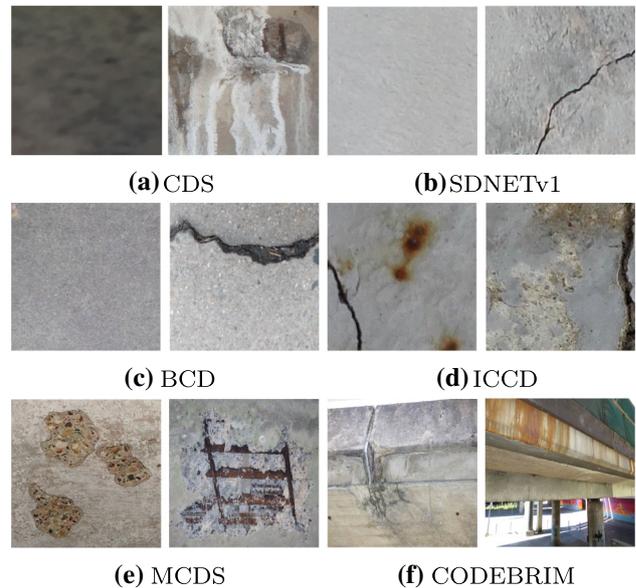

**(a)** CDS  **(b)** SDNETv1  **(c)** BCD  **(d)** ICCD  **(e)** MCDS  **(f)** CODEBRIM

**Fig. 4** Examples from the bridge inspection datasets. From left to right: (**a**) complex lighting with no crack, Crack with efflorescence, and exposed bars. (**b**) Intact concrete, minor crack, (**c**) Intact concrete, large crack on deck, (**d**) crack with rust stains, crack with minor scaling, (**e**) concrete scaling, corrosion with exposed bars, (**f**) spallation and efflorescence, rust stains

Four of the datasets mainly focus on crack detection tasks. The other two datasets are for damage classification in a multi-label setting in which multiple damages coexist on a single input image. We briefly introduce the key characteristics of each dataset.

**CDS** [26]. The Cambridge dataset is a small bridge inspection dataset that seeks to detect concrete defects. The binary classes are divided into *healthy* and *unhealthy* of 691 and 337 images, respectively. Besides different luminous conditions on all images, the unhealthy class consists of concrete damage such as cracking, graffiti, vegetation, and blistering.

**SDNET** [11]. The dataset has 230 images of reinforced concrete decks, walls, and pavements sub-segmented into 56000 images. However, we found misclassified label categories in the datasets. To utilize the data, we performed a manual cleaning of deck images only. The new dataset is referred as *SDNETv1*, which is highly imbalanced having 2, 025 *crack* and 11, 595 *uncracked* images.

**BCD** [67]. Bridge crack detection consists of 4, 058 *crack* and 2, 011 *background* images. The background images consist of healthy concrete surfaces. The dataset contains crack images with details like shading, water stains, and strong lights.

**ICCD** [37]. The image-based concrete crack detection dataset is one of the largest crack datasets having 60,000 images with an equal distribution of *crack* and *uncracked* classes. The images are captured using smartphone cameras under varying lighting conditions. In addition to cracks, the images predominately show corrosion strains.

**MCDS** [28]. The multi-classifier dataset has original inspection data and collected images from ten highway bridges. The authors defined the problem in a multistage classification manner. However, we used the dataset in a multi-label setting having ten classes, i.e., *crack* (789), *efflorescence* (311), *scaling* (168), *spalling* (427), *general defects* (264), *no defects* (452), *exposed reinforcement* (223), *no exposed reinforcement* (203), *rust straining* (355), and *no rust straining* (415).

**CODEBRIM** [42]. The COncrete DEfect BRidge IMage Dataset provides the overlapping defects images of 30 bridges collected via camera and UAVs. The dataset consist of six classes, i.e., *cracks* (2507), *spalling* (1898), *efflorescence* (833), *exposed bars* (1507), *corrosion stain* (1559), and *background* (2506). The authors have framed the damage detection problem as a multi-target multi-class due to the overlapping damages in images. Here, we utilize the dataset in a multi-label setting only.

## 4 Methodology

### 4.1 In-domain and cross-domain transfer strategies

We propose to employ *transfer learning* as a natural solution for learning and improving the performance of damage detection models on small labeled datasets, as shown in numerous other vision tasks [45]. Transfer learning attempts to transfer the learned knowledge from a *source task* to improve the learning in a related *target task* [62]. Here, we consider variants of transfer learning, referred to as *cross-domain* and *in-domain*, and their *combination* as shown in Fig. 5. The main difference between these approaches is the type of dataset used as a source task (also called the upstream dataset).

In *cross-domain transfer learning*, a convolutional neural network is trained over a large IMAGENET dataset having one million images of generic objects [8]. The learned representations are then fine-tuned for the specific downstream tasks, such as for damages detection. Given the computational requirements of network training using the massive IMAGENET dataset, several pre-trained models are made available by academia and industry. The pre-trained models, such as VGG16, Inception-v3, and ResNet50, have shown to yield state-of-the-art performance for object detection and image localization in the ImageNet Large-Scale Visual Recognition Challenge (ILSVRC). The pre-trained models consist of representations (or weights) learned from IMAGENET and a standard CNN architecture.

For *in-domain transfer learning*, the upstream and downstream datasets belong to a similar application domain, for instance, in medical imaging and asset inspection, among others. For comparison purposes, we utilize standard CNN architectures for training on a visual inspection dataset, which are then fine-tuned for a specific task. It is important to note that the dataset used for upstream and downstream tasks are mutually exclusive as depicted by one less square in Fig. 5b. In theory, in-domain transfer learning should provide improved model performance compared to cross-domain due to the availability of similar visual concepts. However, cross-domain transfer learning is standard in the computer vision field as representations are learned from a large dataset having diverse classes.

We also study the *combination of in-domain and cross-domain transfer learning* to enable further performance improvements for damage detection tasks. Here, the pre-trained models having IMAGENET representations are first trained on a visual inspection dataset before fine-tuning to a specific damage detection problem. The combination transfer strategy is depicted in Fig. 5c.

### 4.2 Details of the experiments

We conduct several experiments to evaluate different transfer strategies for improved damage detection tasks. The experiments seek to answer the following questions:

- Do features reused from standard pre-trained models prove helpful compared to learning from scratch for damage detection task?
- Does the standard architecture perform better than small custom CNN models?

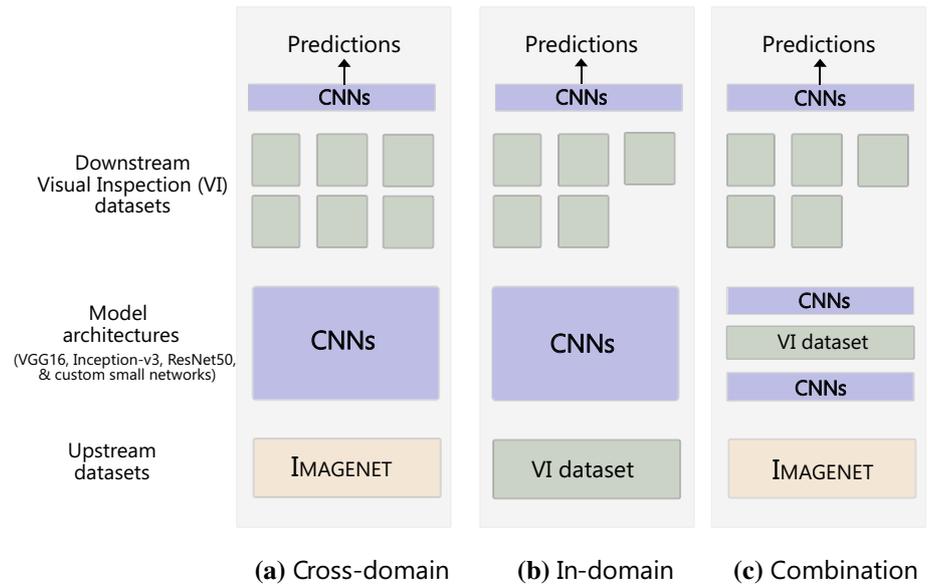

**Fig. 5** Schematic representation of transfer learning strategies

(a) Cross-domain  (b) In-domain  (c) Combination

- Does in-domain transfer learning provide a performance benefit compared to the state-of-the-art cross-domain transfer approach?
- Does the combination of in-domain and cross-domain perform better than their stand-alone versions?
- How well do the representations generalize given an extremely small downstream training dataset?

To acquire pre-trained (IMAGENET) features, we utilize pre-trained models, namely VGG16, Inception-v3, and ResNet50. The choice of these models is motivated by the fact that they are widely used for damage detection tasks, as shown in the literature [6, 28, 34, 68, 77]. The traditional strategies for leveraging transfer learning include: (i) using a pre-trained model as a fixed-feature extractor (FE) in which pre-trained layers are kept frozen, (ii) fine-tuning (FT) all or a few of the layers of an existing model so that the weights are updated for the target task, and (iii) training IMAGENET weights from scratch and then fine-tuning on the target task. Since training on IMAGENET from scratch is computationally expensive, in our experiments, we opted for feature extraction and fine-tuning strategies appended with additional convolution (2D) and a dropout layer to avoid overfitting.

For the comparative evaluation of standard architecture, four small and simpler CNNs models are also adopted from [49]. [49] referred to these models as a *CBR* family and used them to evaluate pre-trained networks for medical imaging. The simpler network consists of four to five layers of convolution, batch normalization, and ReLU activation with varying numbers of filters and kernel sizes. Depending on the architecture configuration, the size of CBR CNNs networks ranges from a third of the IMAGENET architecture size to one-twentieth of the size as described in [49]. Due to the extensive computation costs of training IMAGENET, we perform architectural comparisons with random initialization only.

Besides cross-domain transfer, we investigate in-domain transfer learning for structural damage detection where a dataset from a similar domain is utilized for upstream and downstream tasks. For comparison purposes, we evaluated the best performing random initialized and IMAGENET model for the specific downstream damage detection task. Finally, we conduct several experiments with fewer labeled examples from training sets to assess the quality of learned representation from cross-domain and in-domain learning within a low-data regime.

### 4.3 Evaluation approach and metrics

For the performance evaluation, we employ the held-out approach in which the complete dataset is split into training, validation, and test split with an approximate ratio of 70%, 10%, and 20%. For ICCD and CODEBRIM dataset, the predefined train/validation/test ratios are used. The validation set is used for the hyper-parameter tuning of the model. We perform five independent runs of the models for 30 epochs with a 64 batch size for all the experiments. The learning rate is kept as low as $10^{-4}$ with a binary or categorical cross-entropy loss function depending on the problem. It is important to reiterate that the main focus of this study is to evaluate the network architecture and the source dataset characteristics in the sense of transferring robust representations for structural damage detection. We deliberately avoid the extensive hyper-parameters tuning to compare dataset characteristics and architectural impact on performance. We use the standard parametric configuration of the Keras framework unless explicitly stated above. For the hardware, we use a machine with two Nvidia Tesla T4 GPUs on the Google Cloud Platform.

Additionally, for the performance evaluation and comparison of the deep models, we compute the **AUC-ROC** (area under the receiver operating characteristic curve) score, which is one of the most robust evaluation metrics for classification tasks having skewed class distribution [16]. AUC-ROC computes the area under the curve across all classification thresholds and summarizes the ROC in a single number. We provide a comparison of models using the AUC-ROC score since it is a classification threshold invariant. This means that AUC-ROC score provides information about the model's prediction quality irrespective of the specifically chosen threshold value. The AUC-ROC score ranges in value from 0 to 1. The best performing model will achieve a score closer to 1. A classifier with a zero or 0.5 score is considered random with no predictive capability. For the sake of completeness, we also report accuracy, F-score, precision, and recall scores in Appendix 1.

## 5 Results and comparisons

### 5.1 Transferability of pre-trained models

Five independent runs (i.e., training and evaluation) of CBR models and pre-trained IMAGENET models are performed on six different bridge inspection datasets. The averaged AUC-ROC performance scores and standard deviation are shown in Table 3. First, we note that transfer learning with *fine-tuning* performed consistently well for the majority of bridge inspection datasets, as depicted with bold entries in Table 3.

VGG16 with fine-tuned weights achieved the best performance on all datasets except for CODEBRIM dataset. For BCD, ICCD, and CODEBRIM datasets, the results from Inception-v3 and ResNet50 models are also comparable with the difference of 0.1 scores only. Next, in a *random initialization* setting, the vanilla CBR model, with their one-twentieth to the standard architecture size, performed best on four datasets, followed by VGG16 for the other two datasets as depicted with underlined entries. The relatively sub-optimal performance of standard architectures can be attributed to overfitting due to a heavily parameterized network.

Finally, we note that the *feature extraction* yielded a poor performance compared to random initialization and fine-tuning. This poor performance may be due to considerable differences in the IMAGENET and bridge inspection datasets regarding the number of classes, quality of images, size of the area of interest, and task specification.

### 5.2 Impact of in-domain representation learning

In-domain representation learning typically follows two steps [43]: (a) Learning of representations using in-domain data, called *upstream* training, (b) evaluation of learned representations by transferring them to a new and unseen task, termed as *downstream*. For upstream, we utilize the best performing models that were trained from scratch, and fine-tuned IMAGENET models (see Table 3). The learned representations are further fine-tuned on all the new and unseen downstream tasks.

**Table 3** AUC-ROC performance scores of CBR models and standard IMAGENET architectures with different initializations. VGG16 with fine-tuning performed best for five datasets, followed by ResNet50, as shown with bold entries. Under random initialization and fixed-weight settings, the CBR models performed better than the standard IMAGENET architecture for four datasets followed by VGG16, as shown with underlined entries

|  | CDS | SDNETv1 | BCD | ICCD | MCDS | CODEBRIM |
| --- | --- | --- | --- | --- | --- | --- |
| Random Initialization | | | | | | |
| CBR Tiny | 0.79 ± 0.02 | 0.74 ± 0.01 | 0.97 ± 0.00 | 0.93 ± 0.13 | 0.67 ± 0.02 | 0.80 ± 0.04 |
| CBR Small | 0.78 ± 0.02 | 0.72 ± 0.01 | 0.96 ± 0.0 | 0.98 ± 0.0 | 0.67 ± 0.02 | 0.80 ± 0.03 |
| CBR LargeW | 0.75 ± 0.03 | 0.74 ± 0.01 | 0.97 ± 0.01 | 0.96 ± 0.02 | 0.67 ± 0.01 | 0.80 ± 0.04 |
| CBR LargeT | 0.74 ± 0.02 | 0.67 ± 0.0 | 0.96 ± 0.00 | 0.96 ± 0.02 | 0.64 ± 0.01 | 0.79 ± 0.01 |
| VGG16 | 0.78 ± 0.04 | 0.50 ± 0.00 | 0.98 ± 0.01 | 0.78 ± 0.26 | 0.59 ± 0.01 | 0.82 ± 0.01 |
| Inception-v3 | 0.69 ± 0.03 | 0.60 ± 0.06 | 0.90 ± 0.03 | 0.96 ± 0.01 | 0.61 ± 0.01 | 0.77 ± 0.02 |
| ResNet50 | 0.69 ± 0.02 | 0.61 ± 0.08 | 0.60 ± 0.20 | 0.93 ± 0.03 | 0.57 ± 0.01 | 0.73 ± 0.02 |
| Fine-tuning of pre-trained models. | | | | | | |
| VGG16 | **0.82 ± 0.02** | **0.84 ± 0.0** | **0.99 ± 0.0** | **0.98 ± 0.0** | **0.77 ± 0.03** | 0.88 ± 0.01 |
| Inception-v3 | 0.73 ± 0.02 | 0.78 ± 0.01 | 0.98 ± 0.00 | 0.98 ± 0.0 | 0.72 ± 0.01 | 0.89 ± 0.0 |
| ResNet50 | 0.62 ± 0.02 | 0.82 ± 0.01 | 0.98 ± 0.01 | 0.98 ± 0.0 | 0.54 ± 0.01 | **0.90 ± 0.01** |
| Feature extractions from pre-trained models. | | | | | | |
| VGG16 | 0.77 ± 0.01 | 0.72 ± 0.02 | 0.98 ± 0.01 | 0.93 ± 0.00 | 0.64 ± 0.01 | 0.78 ± 0.00 |
| Inception-v3 | 0.55 ± 0.01 | 0.50 ± 0.01 | 0.52 ± 0.02 | 0.67 ± 0.04 | 0.54 ± 0.0 | 0.55 ± 0.01 |
| ResNet50 | 0.50 ± 0.00 | 0.50 ± 0.00 | 0.50 ± 0.00 | 0.50 ± 0.01 | 0.50 ± 0.00 | 0.50 ± 0.00 |

**Table 4** AUC-ROC performance scores for in-domain transfer per dataset. The rows represent the upstream source model trained for the target task, and the column shows the results of in-domain transfer for each dataset. The bold entries depict the best performing models compared to the random initialization (given in the first row)

|  | **CDS** | **SDNETv1** | **BCD** | **ICCD** | **MCDS** | **CODEBRIM** |
|---|---|---|---|---|---|---|
| **Rand. Init.** | **0.79 ± 0.02** | 0.74 ± 0.01 | 0.98 ± 0.01 | 0.97 ± 0.0 | 0.67 ± 0.02 | **0.82 ± 0.01** |
| **CDS** |  | 0.73 ± 0.05 | 0.99 ± 0.00 | 0.96 ± 0.01 | 0.67 ± 0.02 | 0.80 ± 0.03 |
| **SDNETv1** | 0.75 ± 0.01 |  | 0.98 ± 0.00 | 0.97 ± 0.00 | 0.67 ± 0.01 | 0.77 ± 0.03 |
| **BCD** | 0.77 ± 0.04 | **0.81 ± 0.00** |  | **0.98 ± 0.00** | 0.61 ± 0.01 | 0.81 ± 0.00 |
| **ICCD** | 0.78 ± 0.02 | 0.80 ± 0.00 | **1.00 ± 0.00** |  | 0.69 ± 0.00 | 0.76 ± 0.03 |
| **MCDS** | 0.76 ± 0.02 | 0.72 ± 0.07 | 0.98 ± 0.00 | 0.97 ± 0.01 |  | 0.79 ± 0.02 |
| **CODEBRIM** | 0.75 ± 0.04 | 0.80 ± 0.01 | 0.98 ± 0.00 | 0.98 ± 0.00 | **0.70 ± 0.01** |  |

Upstream model architecture: CBR Tiny (CDS, SDNETV1, MCDS) CBR Small (ICCD)
VGG16 (BCD, CODEBRIM)

Table 4 provides the results of the in-domain transfer. For comparison purposes, the results from random initialization are also provided. The first column depicts the upstream datasets, and the rest of the columns' names are downstream datasets. Out of six datasets, the in-domain transfer yields performance improvement for at least four datasets than learning from scratch. SDNETV1 and MCDS show notable performance improvements with an increase of 0.7 and 0.2 in the AUC-ROC score, respectively. Additionally, BCD and ICCD datasets predominantly showed good transfer of in-domain knowledge across all the target tasks, despite the considerable difference in their sizes.

We also evaluate the *combination of in-domain and cross-domain* transfer for performance gains. Compared to discrete IMAGENET or in-domain transfer, the combination shows further performance improvement for the majority of datasets, as shown with bold entries in Table 5. With the in-domain and cross-domain transfer combination, it is difficult to remark about the usefulness of a specific dataset since the performance scores across the different source datasets are very similar. When used as a source, most of the datasets show comparable performance results irrespective of their size and number of classes. It is essential to mention that the combination of in-domain and cross-domain transfer can introduce additional computation costs due to fine-tuning of large pre-trained models. In the practical setting, the performance gains and computational cost can be compared to gauge the usefulness of this approach.

Figure 6 presents the comparison of loss and performance during training of network with four different initialization settings. These comparisons show the convergence capacity of the network with respect to a specific dataset and initialized weights. It is noted that the network initialization with in-domain and combination of in-domain and cross-domain converges faster than IMAGENET for binary classification datasets, namely for CDS, SDNETv1, and BCD. For multi-label dataset, the IMAGENET network shows better convergence capacity.

**Table 5** AUC-ROC performance scores for in-domain and IMAGENET transfer. The rows represent the upstream model of in-domain and IMAGENET representations, and the column entries show the results of transfer for each dataset. The bold entries depict the best performing models compared to transfer from IMAGENET only (given in the first row)

|  | **CDS** | **SDNETv1** | **BCD** | **ICCD** | **MCDS** | **CODEBRIM** |
|---|---|---|---|---|---|---|
| **IMAGENET** | 0.82 ± 0.02 | 0.84 ± 0.0 | 0.99 ± 0.0 | 0.98 ± 0.0 | 0.77 ± 0.03 | **0.90 ± 0.01** |
| **CDS** |  | 0.84 ± 0.01 | 0.99 ± 0.0 | **0.99 ± 0.0** | 0.75 ± 0.02 | 0.885 ± 0.01 |
| **SDNETv1** | 0.76 ± 0.02 |  | 0.99 ± 0.0 | 0.98 ± 0.0 | 0.70 ± 0.02 | 0.86 ± 0.01 |
| **BCD** | 0.80 ± 0.03 | **0.85 ± 0.0** |  | 0.98 ± 0.0 | 0.75 ± 0.01 | 0.88 ± 0.01 |
| **ICCD** | 0.81 ± 0.02 | 0.84 ± 0.01 | **1.00 ± 0.0** |  | 0.70 ± 0.01 | 0.87 ± 0.01 |
| **MCDS** | **0.84 ± 0.02** | 0.83 ± 0.02 | 0.99 ± 0.0 | 0.98 ± 0.0 |  | 0.88 ± 0.0 |
| **CODEBRIM** | 0.82 ± 0.02 | 0.81 ± 0.01 | 0.99 ± 0.0 | 0.98 ± 0.0 | **0.79 ± 0.0** |  |

Upstream model architecture: VGG16 architecture for all except for CODEBRIM which used ResNet50

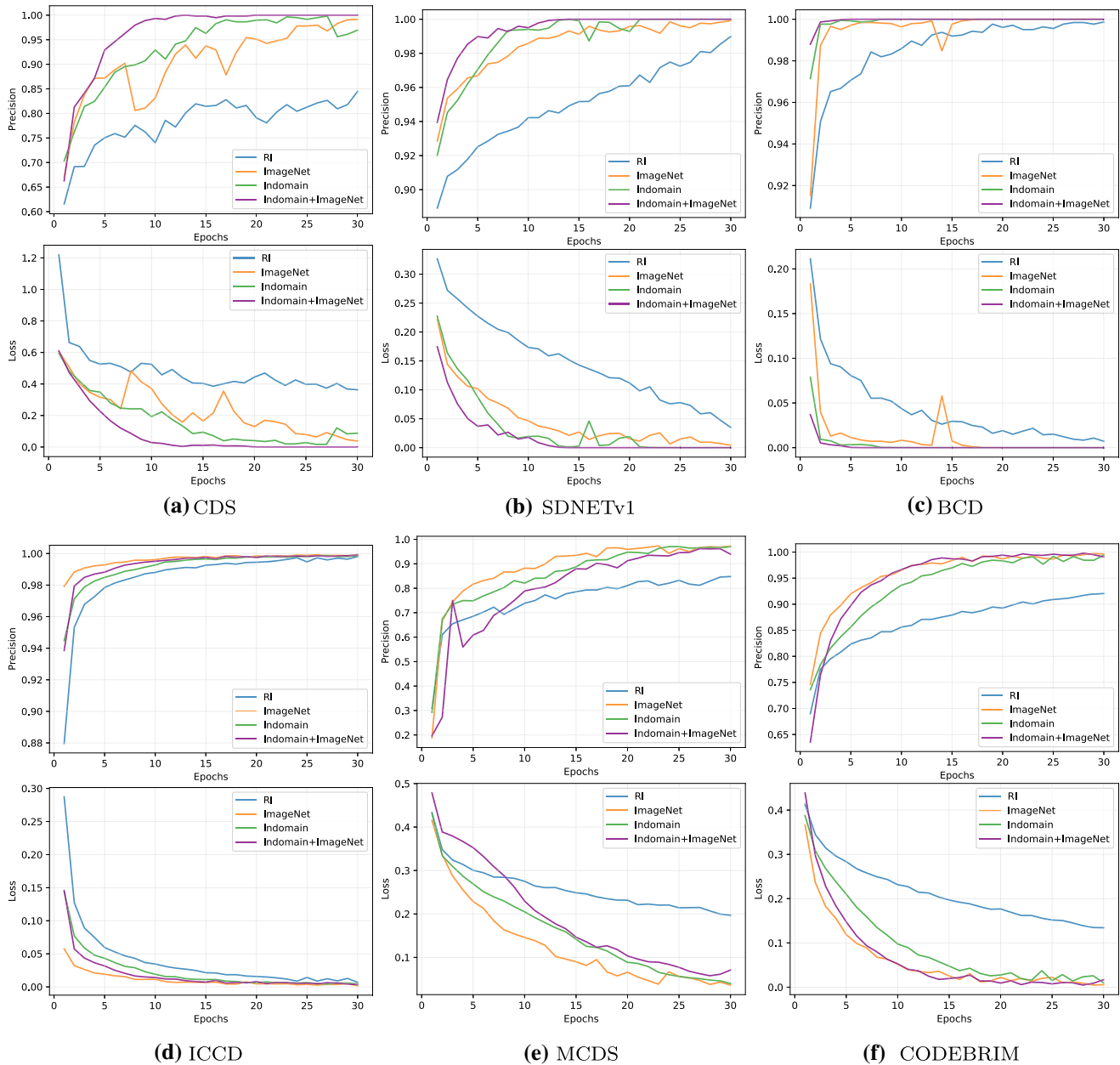

**Fig. 6** Comparison of network training (with precision and loss) for 30 epochs with different initialization settings. RI: random initialization, IMAGENET: pre-trained VGG16 with IMAGENET, in-domain: pre-trained with one of the visual inspection datasets, in-domain+IMAGENET: combination of in-domain and IMAGENET pre-training

## 5.3 Effectiveness of transfer learning in a low-data regime

Bridge inspection datasets are typically smaller than IMAGENET due to the laborious process of acquiring the labeled images. This lack of data necessitates the effective transfer of knowledge from pre-trained models. Thereby, to further explore the generalizability of in-domain and cross-domain representations for a low-data regime, we perform further experiments with a smaller number of training samples ranging from 5%, 10%, 20%, and 50% of the datasets.

Figure 7 reports the AUC-ROC performance score with different initialization settings for fewer training samples of datasets. Transfer learning either with in-domain or cross-domain datasets provides significant performance gains with very small datasets compared to learning from scratch. The combination of in-domain and ImageNet transfer show significant performance with as few as 5% of training samples only.

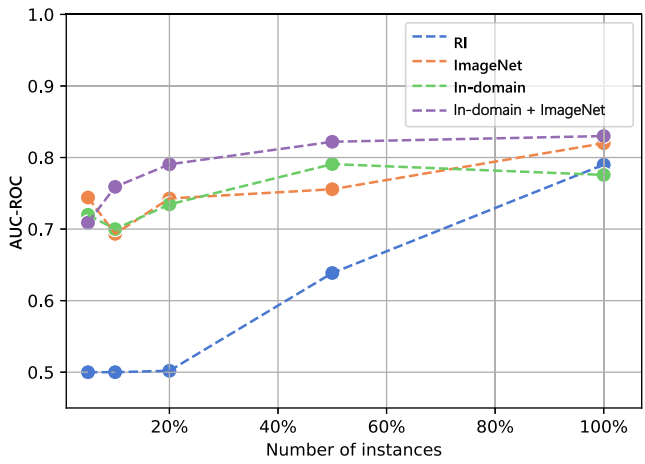
(a) CDS

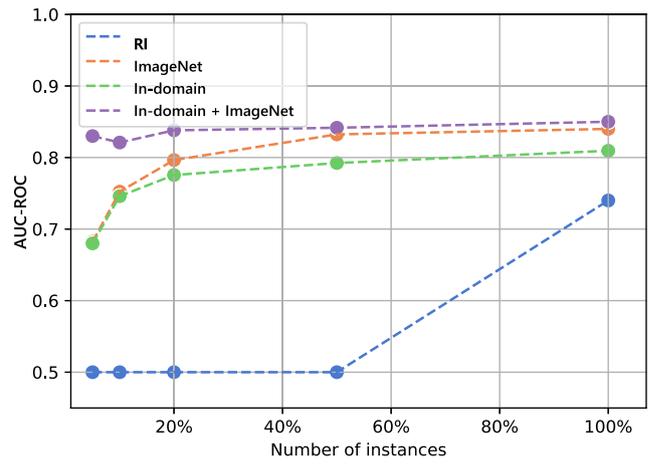
(b) SDNETv1

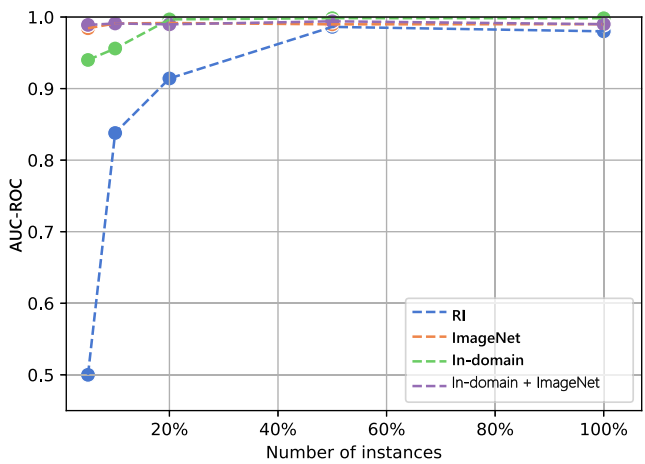
(c) BCD

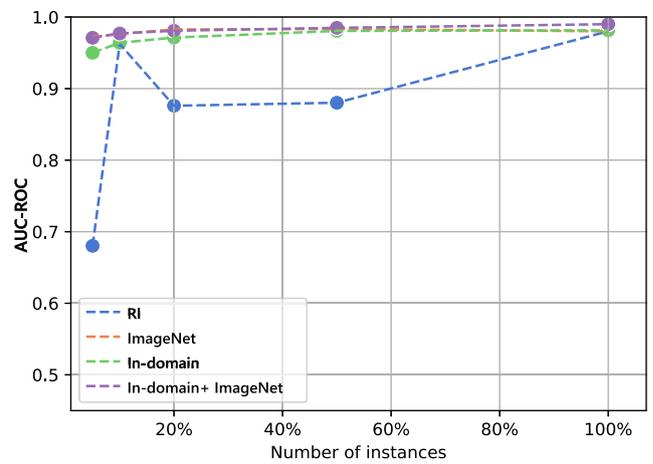
(d) ICCD

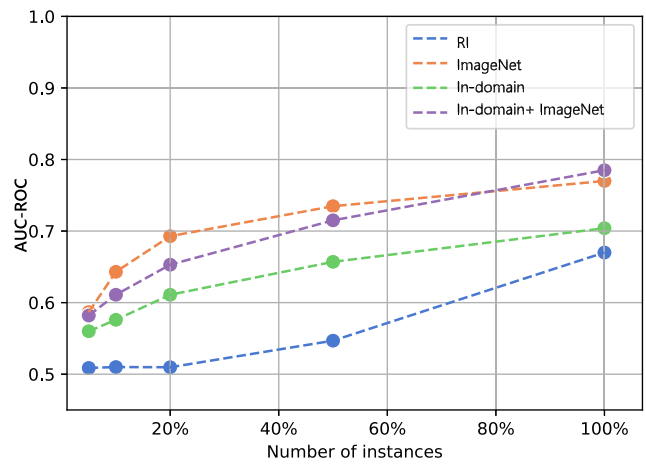
(e) MCDS

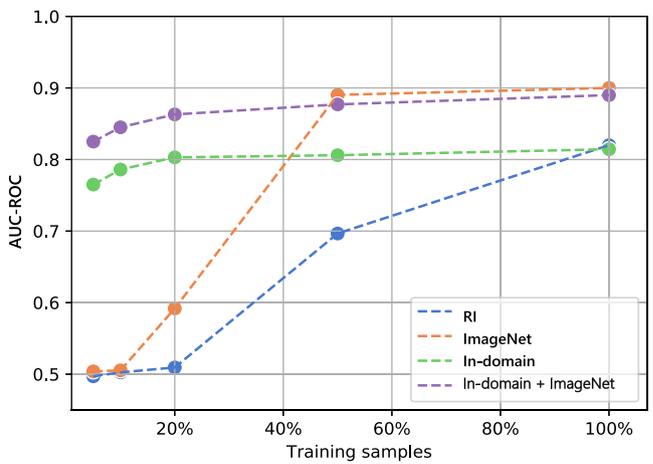
(f) CODEBRIM

**Fig. 7** AUC-ROC score on test set after training with different initialization over a limited number of training samples

Despite significant size differences of the IMAGENET dataset, having millions of images, and the largest in-domain dataset with 60,000 images, their transfer shows comparable performance for binary datasets. For multi-label datasets, i.e., MCDS and CODEBRIM datasets, the impact of different initialization on performance scores is notable. For the CODEBRIM dataset, the in-domain transfer performs notably well compared to IMAGENET when only 5% to 20% data is used. In contrast, with the increasing size of the training samples, the IMAGENET surpasses the other initializations.

To summarize, the combination of in-domain and cross-domain transfer is a desirable transfer setting when the target data is extremely small. However, with larger target datasets, the combination of in-domain and cross-domain transfer yield similar performance as when only IMAGENET is used as source dataset.

### 5.4 Comparison with baselines

This section conducts a comparison to the baselines to assess the performance of the proposed transfer learning strategies. Table 6 presents the accuracy scores for each dataset and provides the comparison to the baselines. Transfer learning with a combination of in-domain and cross-domain strategies shows improved results for at least four datasets as depicted by bold entries. For the other two datasets, i.e., [42] and [28], a direct comparison is not feasible due to the disparity in research methodologies. [42] treated the damage classification problem in multi-label multi-target settings in which the exact match of predicted and actual multi-labels is ensured. [28] dealt with damage classification in a multistage multi-classification manner, where several multi-class and binary classifier are defined. Instead, in our approach, we develop a single classifier for multiple damages detection. Besides transfer learning, the results provided in Table 6 can be used for future performance comparisons and improvements for the publicly available visual inspection datasets.

It is worth noting that different IMAGENET architectures and transfer learning strategies introduced varying computational costs. Pre-trained IMAGENET and in-domain models can be efficient for feature extraction and fine-tuning. The combination of cross-domain and in-domain models can be computationally expensive. However, the combination of cross and in-domain transfer can provide superior performance when the target dataset is extremely small.

## 6 Interpretability of damage detection models

Transparency and interpretability of predictive models are indispensable to enable their use in practice. Deep neural networks are well known for exploiting millions of parameters, processed by several nonlinear functions and pooling layers to learn optimal weights. It is intractable for humans to follow the exact mapping of data from input to classification through these complex multilayered networks. To encourage CNN's usage in practice, the interpretability and explainability of deep neural networks are increasingly popular and an active research area. Several model-agnostic, visual explanations and example-based methods have recently been proposed [15, 41].

In this study, we employed gradient-weighted class activation mapping (Grad-CAM) to visualize and localize the areas of an input image that are most important for the models' prediction. Grad-CAM is a class-discriminative localization technique that, unlike its predecessor [76], does not require any change in the CNNs architecture or retraining for generating visual explanations [54]. Several studies have shown that the deeper layers of CNNs capture high-level abstract constructs; therefore, Grad-CAM utilizes the gradient information flowing into the last convolution layer to evaluate each neuron's importance. We apply importance scores with logistic regression and a nonlinear activation function to obtain coarse heatmaps for discriminative class mappings.

Figure 8 shows the visual explanation of the fine-tuned VGG16 model, which has predominantly performed well for most of the damage detection datasets (see Table 3 for results). We utilized the tf-explain implementation of Grad-CAM to generate visual mapping [51]. In addition to

Table 6 Accuracy scores of transfer learning strategies and comparisons with baselines. The bold entries depict the best performing transfer learning strategy

|  | CDS (%) | SDNETv1 (%) | BCD (%) | ICCD (%) | MCDS (%) | CODEBRIM (%) |
| --- | --- | --- | --- | --- | --- | --- |
| Rand. Init. | 82 | 94 | 98 | 98 | 68 | 82 |
| Cross-domain | 86 | 95 | 98 | 98 | 76 | **90** |
| In-domain | 83 | 95 | **99** | 98 | 69 | 82 |
| Combination | **87** | **96** | **99** | **99** | 79 | **90** |
| Previous studies | – | 92 [11] | 96 [67] | 99 [37] | 85* [28] | 72** [42] |

*Multiple (binary and multi-class) classifiers approach

**Multi-target accuracy score

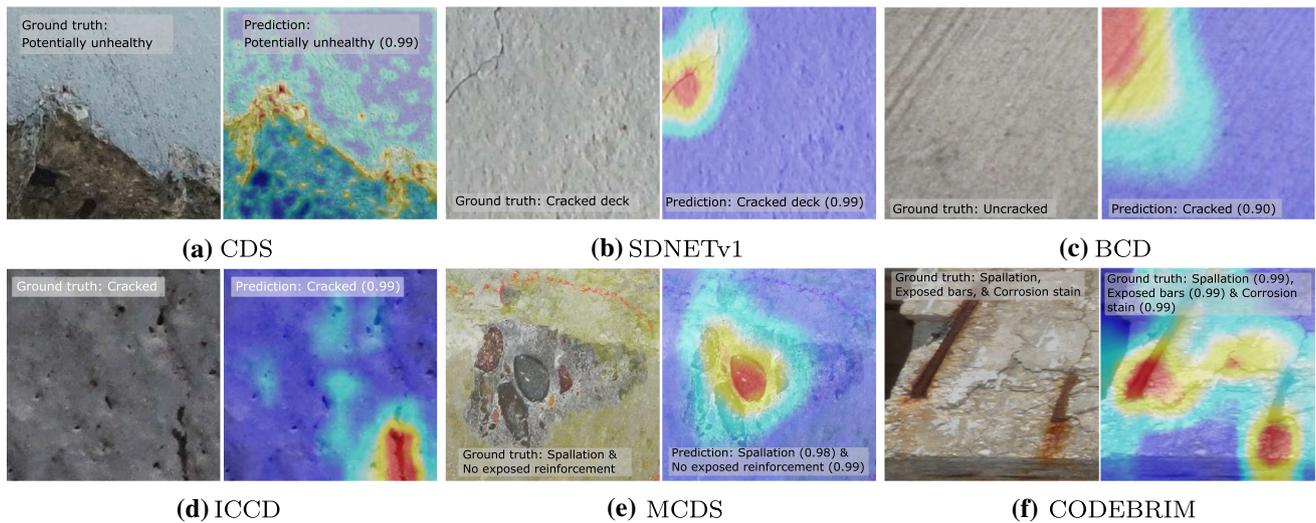

**Fig. 8** Visual explanations for classification and localization of VGG16 models for all six datasets. Grad-CAM highlights class discriminative features. The heatmap localizes the specific damage regions

the last convolution layer of CNN, Grad-CAM can be used to extract discriminative class mapping with any intermediate convolution layer. We experimented with various convolution and pooling layers to generate visual explanations, as shown in Fig. 8. The visual explanations with the correct predicted class validate that the model is paying attention to the right areas of an image to identify and localize a specific damage category. It is interesting to notice Fig. 8, where the model misclassified the healthy patch of concrete as cracked by assigning higher weights to the textured concrete patch.

The visual interpretations of CNN models reveal *why the models predict, what they predict*. The visual explanation also provides localization of specific damage in an input image without additional labeling or segmentation activity. By revealing the decision logic of these complex models, the decision-makers and infrastructure managers can trust them for automatic damage detection tasks.

# 7 Conclusions

This paper presents transfer learning strategies and conducts comprehensive experiments to evaluate their usefulness for damage detection tasks. We compared different initialization settings, namely random initialization, in-domain, cross-domain, and their combination for transfer learning on six publicly available visual inspection datasets of concrete bridges. We found that cross-domain transfer yields performance improvements only when fine-tuned for the target damage detection task.

Our main message is that the combination of in-domain and cross-domain representations provide enhanced performance compared to their stand-alone versions. Additionally, in contrast to pre-trained IMAGENET models, in-domain transfer provides training efficiency and flexibility in selecting relatively smaller yet powerful CNNs architecture. In our exploration of learning for tasks with a limited number of training samples, in-domain and IMAGENET representations show comparable performance. The results demonstrate considerable performance gains when in-domain and cross-domain (IMAGENET) representations are used jointly for the target dataset having fewer training samples.

We further assessed the best performing IMAGENET models by developing visual explanations using gradient-weighted class activation mapping. The visual explanation shows the class discriminative regions of an input that are most paramount for specific damage prediction. Additionally, the interpretability of models' prediction localizes the specific damages and enables decision-makers to understand the underlying intrinsic decision logic of the neural model. Such visual exploration of damage detection and localization also encourages the use of predictive models in practice.

# Additional results

We provide additional results to compare different transfer learning strategies in Table 7. In addition to AUC-ROC scores, we provide accuracy, F-score, precision, and recall measures. *Accuracy* is essentially computed by taking a fraction of correctly predicted samples to the total number of samples. However, in the case of an imbalanced dataset, *accuracy* as performance metric can be misleading.

**Table 7** Performance evaluation of transfer learning strategies with different initialization

|  |  | CDS | SDNETv1 | BCD | ICCD | MCDS | CODEBRIM |
|---|---|---|---|---|---|---|---|
| RI | Acc. | 0.804 | 0.890 | 0.976 | 0.788 | 0.589 | 0.821 |
|  | F-score | 0.802 | 0.838 | 0.977 | 0.721 | 0.229 | 0.724 |
|  | Precision | 0.812 | 0.791 | 0.981 | 0.688 | 0.194 | 0.741 |
|  | Recall | 0.804 | 0.890 | 0.976 | 0.788 | 0.375 | 0.745 |
| Cross-domain | Acc. | 0.864 | 0.955 | 0.986 | 0.985 | 0.780 | 0.900 |
|  | F-score | 0.860 | 0.952 | 0.986 | 0.984 | 0.629 | 0.847 |
|  | Precision | 0.865 | 0.953 | 0.987 | 0.985 | 0.697 | 0.859 |
|  | Recall | 0.864 | 0.955 | 0.986 | 0.985 | 0.638 | 0.850 |
| In-domain | Acc. | 0.711 | 0.948 | 0.975 | 0.980 | 0.690 | 0.800 |
|  | F-score | 0.707 | 0.944 | 0.976 | 0.980 | 0.465 | 0.681 |
|  | Precision | 0.708 | 0.945 | 0.979 | 0.980 | 0.579 | 0.751 |
|  | Recall | 0.711 | 0.948 | 0.975 | 0.980 | 0.488 | 0.704 |
| Combination | Acc. | 0.869 | 0.955 | 0.989 | 0.986 | 0.785 | 0.885 |
|  | F-score | 0.864 | 0.953 | 0.989 | 0.986 | 0.642 | 0.822 |
|  | Precision | 0.871 | 0.953 | 0.990 | 0.986 | 0.680 | 0.825 |
|  | Recall | 0.869 | 0.955 | 0.989 | 0.986 | 0.623 | 0.834 |

Therefore, we report weighted accuracy and weighted F-score here. The *weighted* aspect considers the data imbalance of the target class and assigns the weights to data samples accordingly. The *F-score* is a combination of precision (or positive predictive value) and recall (sensitivity) measures. The precision determines the exactness of the model, whereas the recall provides a measure of the model's completeness.

**Author Contributions** All authors contributed equally to the study conception and design. Material preparation, data collection, and analysis were performed by first author. The draft of the manuscript was written by first author and all authors reviewed and corrected multiple versions of the manuscript. All authors read and approved the final manuscript.

**Funding** This research has been partially funded by NWO under the grant PrimaVera NWA.1160.18.238.

## Declarations